%% file: root.tex
\documentclass[letterpaper, 10 pt, conference]{ieeeconf}
\IEEEoverridecommandlockouts
\usepackage{graphicx}
\usepackage{caption}
\usepackage{subcaption}
\usepackage{float} 
\usepackage{xcolor}
\usepackage{cite} 
\usepackage{amsmath,amssymb,amsfonts}
\usepackage{multirow}
\usepackage{physics}
\usepackage{comment}

\usepackage{hyperref}
\usepackage{cancel}
\hypersetup{
    colorlinks=true,
    linkcolor=blue,
    filecolor=magenta,      
    urlcolor=cyan,
}
\begin{document}
\title{SeanNet: Semantic Understanding Network for Localization Under Object Dynamics

\thanks{Xiao Li, Yidong Du, Odest Chadwicke Jenkins are with the University of Michigan, Ann Arbor, MI 48109, USA. hsiaoli, duyidong, ocj@umich.edu. Zhen Zeng is with J.P.Morgan AI Research, Ann Arbor, MI 48109, USA.  zhen.zeng@jpmchase.com *This work was done individually from J.P.Morgan AI Research.}

}

\author{Xiao Li, Yidong Du, Zhen Zeng, Odest Chadwicke Jenkins}

\maketitle


\begin{abstract}
    \input{tex/abstract}
\end{abstract}


\section{Introduction}\label{sec:intro}
\input{tex/intro}
\section{Related Work}\label{sec:relatedWork}
\input{tex/relatedWork}
\section{Problem Statement}\label{sec:problem}
\input{tex/problem}

\section{Scene Embedding with SeanNet Model}\label{sec:method}
\input{tex/method}

\section{Experimental Results}\label{sec:results}
\input{tex/results}
\section{Visual Navigation Application}\label{sec:application}
\input{tex/application}
\section{Conclusion}\label{sec:conclusion}
\input{tex/conclusion}
\bibliographystyle{IEEEtran}
\bibliography{root.bib} 
\end{document}

%% file: tex/abstract.tex
We aim for domestic robots to perform long-term indoor service. Under the object-level scene dynamics induced by daily human activities, a robot needs to robustly localize itself in the environment subject to scene uncertainties. Previous works have addressed visual-based localization in static environments, yet the object-level scene dynamics challenge existing methods for the long-term deployment of the robot. This paper proposes a SEmantic understANding Network (SeanNet) architecture that enables an effective learning process with coupled visual and semantic inputs. With a dataset that contains object dynamics, we propose a cascaded contrastive learning scheme to train the SeanNet for learning a vector scene embedding. Subsequently, we can measure the similarity between the current observed scene and the target scene, whereby enables robust localization under object-level dynamics. In our experiments, we benchmark SeanNet against state-of-the-art image-encoding networks (baselines) on scene similarity measures. The SeanNet architecture with the proposed training method can achieve an 85.02\% accuracy which is higher than baselines. We further integrate the SeanNet and the other networks as the localizers into a visual navigation application. We demonstrate that SeanNet achieves higher success rates compared to the baselines.

%% file: tex/intro.tex
Domestic robots are expected to perform long-term service in changing environments where objects often move around because of human activities. Robots should act robustly under scene uncertainties due to such object-level dynamics where localization is crucial. With recent advances in deep learning and computer vision, deep neural networks enable end-to-end visual-based localization and navigation. However, existing works~\cite{SPTM, chaplot2020neural} assumed static environments. Visual-based localization under scene uncertainties due to object-level dynamics is still an open challenge that remained to be addressed. 

\begin{figure}[!t]
\begin{center}
\includegraphics[width=1\linewidth]{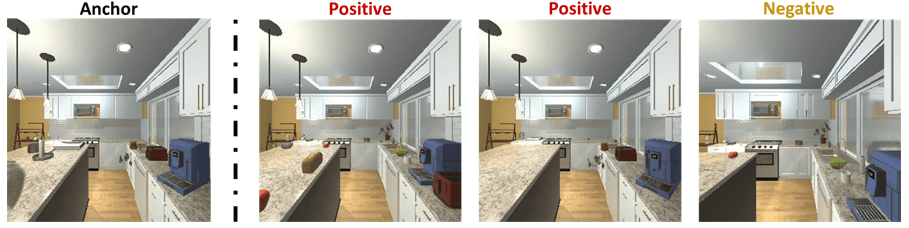}
\end{center}
\caption{Data in our triplet dataset: given a target anchor image, reasoning from positions of the static objects (e.g. counter-Top and cabinets) and filtering out the noise of the dynamic objects (e.g. coffee machine and toaster), one can conclude that the positive images are observed from the same pose as the anchor one while the negative image is less similar to the anchor image due to robot pose deviation.}
\label{fig:triplet_example}
\vspace{-1.5em}
\end{figure}

Consider an example in Fig.~\ref{fig:triplet_example}, the changes between the anchor and positive scenes showcase typical object dynamics observed in indoor environments. Existing localization networks, which purely relying on visual aspects of scenes, hardly learn what visual changes are allowed (e.g. anchor vs. positive) and not allowed (e.g. anchor vs. negative) for matching robot observation to an anchor scene, as we show later in experiments. Our work combines both visual and semantic aspects to reliably match robot observations to anchor scenes for visual localization under uncertainty.

This paper addresses robust visual-based localization under object-level dynamics in indoor scenes. The goal is to enable the robot to learn that the positive and anchor images in Fig.~\ref{fig:triplet_example} correspond to the same robot pose under object dynamics while the anchor and negative ones are not. To achieve this, we propose a network (SeanNet) that can comprehend robot observation from both visual and semantic perspectives. The network architecture integrates a graphical convolution network~\cite{gcn} (GCN) with a convolution neural network (CNN) to embed an observed scene as a vector. The GCN aggregates vector embeddings of the objects' visual features, class labels, and detection bounding boxes guided by the scene graph. The similarity between scenes is measured as cosine similarities of the embedded vectors. SeanNet robustly matches anchor-positive scenes under object dynamics, while still being able to distinguish negative-anchor scenes given ambiguity, therefore, enabling robust similarity-based visual localization.

We use the AI2THOR~\cite{ai2thor} simulation environment to validate the robustness of the SeanNet in similarity measure, where the robot needs to identify positive scenes out of anchor-positive-negative $(A,P,N)$ scene triplets, compared to the state-of-the-art image encoding networks. The performance of the SeanNet is further demonstrated with real-world examples. Moreover, we include a visual navigation application to demonstrate one of the application scenarios for similarity-based localization. Our main contributions as as follows: (1) We propose the SeanNet architecture that learns a vector scene embedding in both visual and semantic aspects. (2) We introduce a cascaded contrastive learning scheme with a dataset containing object dynamics to facilitate the training of the SeanNet for similarity measure. (3) We show SeanNet provides a robust similarity measure and localization performance under scene uncertainty.

\begin{figure*}[t!]
\vspace{0.4em}
\begin{center}
\includegraphics[width=1\textwidth]{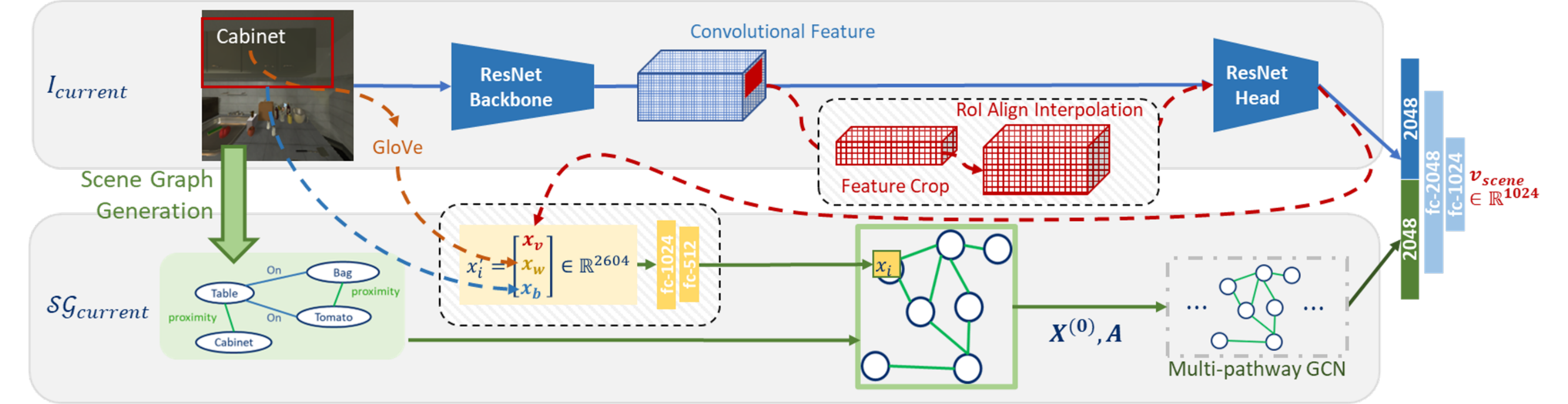}
\end{center}
\vspace{-1em}
\caption{\textbf{Overview of SeanNet architecture.} The scene graph branch takes learned object embedding vectors as node features and learns a comprehensive scene graph embedding while the image branch is responsible for learning a  visual embedding. Visual and scene graph embedding vectors are fused to a single scene embedding vector as the output of SeanNet.}
\label{fig:seanNet}
\vspace{-1em}
\end{figure*}

%% file: tex/relatedWork.tex
\textbf{Learning a similarity measure}\quad Human learns similarity and difference between instances more effectively by comparison. Schultz and Joachims~\cite{comparativeLearning} first transfer the pairwise similarity measurement task to the problem of relative comparison within triplets. The survey from~\cite{surveyMetric} provides an overview of metric learning strategy while Hoffer and Ailon~\cite{hoffer2015deepTriplet} give a comprehensive discussion in similarity measurement. Grounded on the philosophy of triplets, people establish CNN-based methods that measure similarity between human faces subject to the viewpoint and light variation~\cite{facenet, defenseTriplet}. Inspired by triplet loss~\cite{facenet}, our work develops and trains a network that measures 
scene similarity in indoor scenes with object dynamics.  

\textbf{Localization under dynamics}\quad Given current and target images, Fraundorfe et al.~\cite{fraundorfer2007topological} localize the robot to target pose by comparing SIFT features~\cite{SIFT} and Savinov et al.~\cite{SPTM} train a Siamese network for binary localization decision. Meanwhile, there is significant progress in outdoor visual place recognition~\cite{VPR} (VPR) under changing perceptual conditions~\cite{naseer2017semantics, pirker2011cd, NetVLAD}. In contrast, our network implicitly learns to reasoning the object-level dynamics guided by the network architecture. Our method combine both visual and semantic features to learn the appearance and the structure of the scene which enables a robust similarity-based localization under object dynamics. People have also explored the potential of actively tracking~\cite{GATMO, DATMO}, modeling and inferring indoor object dynamics~\cite{pulido2016persistent, bore2018multiple, zeng2020semantic}, but they assume the robot being well-localized.

%% file: tex/problem.tex
We consider an indoor environment with a set of common household objects $\{o_i\}_{i=1}^{M}$, where $M$ is the maximum number of instances in the environment. The object dynamics induce scene uncertainty, where the other factors such as lighting variation is not in the scope of our problem setup. We consider a robot lives in the environment with its pose $[p_x,\;p_y,\;p_{\theta}]^T\in\mathbb{R}^3$ in a reachable space $\mathcal{P}$. At an arbitrary pose $A\in\mathcal{P}$, the robot can observe feature $f_A$. For poses $A,B\in\mathcal{P}$, we aim to measure the closeness between $A$ and $B$ in space $\mathcal{P}$ under similarity metric $\mathcal{S}$ where $\mathcal{S}(f_A,f_B)\in[0,1]$.

%% file: tex/method.tex
\begin{figure*}
\vspace{0.4em}
\begin{center}
\includegraphics[width=1\textwidth]{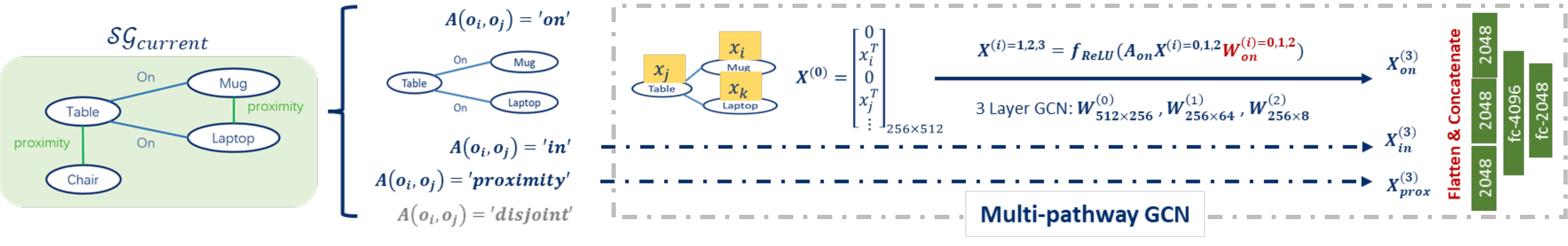}
\end{center}
\vspace{-1em}
\caption{\textbf{Architecture of multi-pathway GCN.} We decompose the adjacency matrix $A$ into four individual adjacency matrices by the inter-object relationship edge types. Except for the disjoint relationship, we propagate the other three adjacency matrices together with the initial feature matrix $X^{(0)}$ through their designated GCN pathways and generate three 2048-d vectors. The final fcn concatenates the three vectors and generates a 2048-d scene graph embedding.}
\label{fig:multigcn}
\vspace{-1em}
\end{figure*}

The SeanNet (\ref{subsec:seannet_arch}) can comprehend the object dynamics by aggregating both visual and semantic information in object embeddings (\ref{subsec:objEmbedding}) through a multi-pathway GCN (\ref{subsec:multigcn}). Subsequently, with a cascaded constrastive learning scheme (\ref{subsec:dataset}), the SeanNet learns a scene embedding for measuring similarity between the scenes, whereby we can develop a similarity-based localization (\ref{subsec:similarityMeasure+localization}) that is robust to scene uncertainties due to object dynamics. 

\subsection{SeanNet Architecture}\label{subsec:seannet_arch}
The network architecture is based on ResNet50~\cite{resnet} and GCN~\cite{gcn}. As shown in Fig.~\ref{fig:seanNet}, SeanNet comprises two branches and takes the current feature $f_{\text{current}}=\{I, \mathcal{SG}\}$ as input where $I$ is an RGB image and $\mathcal{SG}$ is a scene graph. Each scene graph $\mathcal{SG}(o,R)$ is a graphical decomposition of $I$ that comprises objects $\{o_i\}\subset \{o_i\}_{i=1}^{M}$ as nodes and the relationships $R_{ij}\in\mathcal{R}=\left\{\text{in},\;\text{on},\;\text{proximity},\;\text{disjoint}\right\}$ between $o_i$ and $o_j$ as edges. Each $o_i$ is detected from $I$ with a semantic class label and a 2-D bounding box. 

\noindent\textbf{Image Branch}\quad It takes a $224\times224$ RGB image and generates a 2048-d vector. We decompose the ResNet-50 into the backbone (ResNet-50-C4, 1-4th stage) and the head (5th stage without final full-connected layer fc-1000) where the backbone passes a $14\times14\times1024$ convolutional feature to the head. Meanwhile, together with the object detection labels and bounding boxes, the image branch passes object visual feature embeddings (\ref{subsubsec:x_v}) to scene graph branch for learning vector embeddings as objects' features (\ref{subsec:objEmbedding}). 

\noindent\textbf{Scene Graph Branch}\quad Each $\mathcal{SG}$ can be uniquely represented by an adjacency matrix $A$. Each non-empty entry $A(o_i,o_j)$ in the $i^{th}$ row and $j^{th}$ column of matrix $A$ is associated with a inter-object relationship $R_{ij}$. For instance, $A(o_i,o_j)=\text{'on'}$ means $o_j$ on $o_i$. The scene graph branch takes a $M\times M$ adjacency matrix $A$ and a $M\times V$ feature matrix $X^{(0)}$ and generates a 2048-d vector through a multi-pathway GCN as in~\ref{subsec:multigcn}. The $i^{th}$ row of $X^{(0)}$ is a learned object feature embedding $x_i$ for a detected object $o_i$ (\ref{subsec:objEmbedding}). 

\subsection{Object Feature Embedding}\label{subsec:objEmbedding}
We aim to integrate the visual and semantic information and learn a vector representation for a detected object. As shown in Fig.~\ref{fig:seanNet}, we concatenates vectors $x_b,\;x_v,\;x_w$ which are positional feature, visual feature and word embeddings. We passes them through a fully connected network (fcn) which generates $x_i$ as the embedding of detected object $o_i$. The three embedding vectors guarantee the network captures different aspects of information in the learning process.

\subsubsection{Positional feature embedding}\label{subsubsec:x_b}
For a detected object $o_i$ in image, the 256-d vector $x_b$ is the embedding of the bounding box as the positional feature. One can uniquely represent a bounding box of $o_i$ by the lower-left and upper-right corner pixels $(u_1,v_1)$ and $(u_2,v_2)$. We use position embedding function $PE(\cdot):\mathbb{R}\rightarrow \mathbb{R}^{64}$ from~\cite{attention} and transfer each coordinate in the pixel to a 64-d vector. Then, the bounding box embedding of $o_i$ can be represented as 
\begin{equation*}
    x_b=\left[PE(u_1)^T\;PE(v_1)^T\;PE(u_2)^T\;PE(v_2)^T\right]^T\in\mathbb{R}^{256}.
\end{equation*}
The sinusoidal function nature of the $PE(\cdot)$ guarantees a similar vector embedding for bounding boxes if their corresponding corner pixels are close to each other. 

\subsubsection{Visual feature embedding}\label{subsubsec:x_v}
For a detected object $o_i$, we can crop a regional convolutional feature from the output of the ResNet backbone by the relative bounding box coordinates. We apply RoI align~\cite{roiAlign} and interpolate a full-dimension convolutional feature of $14\times14\times1024$. Afterward, SeanNet passes the full-dimension convolutional feature to the ResNet head which generates a 2048-d embedding vector as the object's visual feature $x_v$. We use the visual feature to capture the state of the object such as pose and texture which might not be presented in $x_b$.

\subsubsection{Word embedding}\label{subsubsec:x_w}
We use GloVe word embedding~\cite{glove} to translate the object semantic label (e.g. "mug") to a 300-d vector $x_w$ which introduces semantic understanding and general knowledge of the objects to the network.

\subsection{Multi-pathway GCN}\label{subsec:multigcn}
In a scene graph $\mathcal{SG}(o,R)$, each node represents an object $o_i$ with its unique feature embedding $x_i$. Moreover, the scene graph as an entity encodes semantic information regarding inter-object interaction. We leverage the GCN structure and propagate the learned object feature embedding through the scene graph edges $R_{ij}$. Eventually, SeanNet integrates the object feature embedding in $X^{(0)}$ with the relationship information in $A$ through a multi-pathway GCN and generates a comprehensive vector embedding for a scene graph.

We develop a multi-pathway GCN to propagate the information in the scene graph. The information propagation in different inter-object relationships $R_{ij}\in\mathcal{R}$ are supposed to be distinctive. Thus, we decompose the adjacency matrix $A$ as $A=A_{\text{on}}+A_{\text{in}}+A_{\text{proximity}}+A_{\text{disjoint}}$ where each single decomposed $A_{r}$ encodes a subset of scene graph with single edge type $r\in \mathcal{R}$. Moreover, the non-informative 'disjoint' relationship is not useful in information propagation.

SeanNet passes three matrices $A_{\text{on}},\;A_{\text{in}},\;A_{\text{proximity}}$ separately with feature matrix $X^{(0)}$ through different GCN pathways as in Fig. ~\ref{fig:multigcn}. Each pathway contains three-layer GCN. Each layer propagates the information between connected objects and downsize the feature matrix as 
\begin{equation*}
    X_r^{(i)}=f_{ReLU}(\hat{A_r}X_r^{(i-1)}W^{(i)}),\; i=1,2,3
\end{equation*}
where $\hat{A_r}$ is the row-normalized~\cite{gcn} adjacency matrix and $W^{(i)}$ is the parameter of $i^{th}$ GCN layer. At the end, we have three feature matrices $X_{\text{on}},\;X_{\text{in}},\;X_{\text{proximity}}$ of $256\times 8$ which are further flattened and concatenated to a 6144-d vector.

\subsection{Similarity Measure and Localization}\label{subsec:similarityMeasure+localization}
With SeanNet architecture, we can build a Siamese network $\mathcal{L}(f_A, f_B)$  for similarity measure. The network $\mathcal{L}$ takes two input features $(f_A, f_B)$. Features are first processed by SeanNet separately and obtain 2048-d vectors $(v_A, v_B)$. Afterward, the network $\mathcal{L}$ calculates the cosine projection as similarity measurement 
\begin{equation*}
    \mathcal{S}(f_A,f_B) = \mathcal{L}(f_A, f_B) = \frac{v_A\cdot v_B}{\norm{v_A}_2\norm{v_B}_2}\in \mathbb{R}.
\end{equation*}
Then, we can choose a threshold $\epsilon$. $\mathcal{L}(f_A, f_B)>\epsilon$ implies the current feature obtained at pose $B$ is sufficiently similar with that from $A$ such that $f_A, f_B$ are observations taken at the same robot pose.

\subsection{Cascaded Contrastive Learning Scheme}\label{subsec:dataset}
To simplify the problem, we define $\mathcal{P}=\left\{[i\cdot d,\;j\cdot d,\;\frac{\pi\cdot k}{2}]^T\in\mathbb{R}^3|\; i,j,k\in \mathbb{Z}\right\}$ as a grid with cell length $d$, and the robot has four possible headings. We use a triplet dataset akin~\cite{facenet} containing object dynamics in our network training to avoid manually setting ground-truth similarity labels. Each data point in our dataset is a triplet $(f_A,f_P,f_N)$ where the anchor feature $f_A$ is more similar to the positive feature $f_P$ than to the negative one $f_N$. Then, we employ the triplet loss function to guarantee such a property during training
\begin{equation*}
    \ell(f_A,f_P,f_N) = \max\{0,\; \mathcal{L}(f_A, f_N) + \alpha - \mathcal{L}(f_A, f_P)\}
\end{equation*}
where $\alpha$ is a constant margin to ensure $\mathcal{L}(f_A, f_P) - \mathcal{L}(f_A, f_N) > \alpha$.

The triplet dataset is constructed in a cascaded manner as in Table~\ref{tab:dataset} where the negative data has a larger Manhattan distance deviation from the anchor compared to the positive one. To enforce the learning of localization against object dynamics, $50\%$ positives and anchors are collected at the same robot poses with different object layouts as in Fig.~\ref{fig:triplet_example} while the negative data can be obtained one grid cell $d$ away from the anchor. Notably, the constant margin $\alpha$ and the cascaded triplet dataset collectively ensure a prominent threshold $\epsilon$ as an empirical localization decision parameter. 

\begin{table}[H]
\vspace{-0.5em}
\centering
\caption{\textbf{Cascaded triplet dataset composition.} The positive and negative data are collected at points $[p_x+\alpha d,\;p_y+\beta d,\;p_{\theta} +\gamma \cdot \frac{\pi}{2}]^T$ which are deviated from the anchor location $[p_x,\;p_y,\;p_{\theta}]^T$.}
\begin{tabular}{ c|c|c } 
\hline\hline
Percen- &  \multicolumn{2}{c}{Deviation $(\alpha, \beta, \gamma)\in\mathbb{Z}^3$ from anchor}\\\cline{2-3}
 tage (\%) &  Positive & Negative \\\hline\hline
50 & $\alpha=\beta=\gamma=0$ & $\abs{\alpha}+\abs{\beta}=1,\;\gamma=0$\\ 
20 & $\abs{\alpha}+\abs{\beta}=1,\;\gamma=0$ & $\abs{\alpha}+\abs{\beta}=2,\;\gamma=0$\\ 
15 & $\abs{\alpha}+\abs{\beta}=2,\;\gamma=0$ & $\abs{\alpha}+\abs{\beta}=3,\;\gamma=0$\\ 
10 & $\abs{\alpha}+\abs{\beta}=3,\;\gamma=0$ & $\abs{\alpha}+\abs{\beta}=4,\;\gamma=0$\\
5 & $\abs{\alpha}+\abs{\beta}=4,\;\gamma=0$ & $\gamma\neq0$\\ 
\hline\hline
\end{tabular}
\label{tab:dataset}
\end{table}

%% file: tex/results.tex
We use AI2Thor~\cite{ai2thor} as the simulation environment and provide a detailed experiment setup in \ref{subsec:Environment}. The training hyper-parameter is provided in \ref{subsec:hyperparam}. We evaluate our SeanNet architecture in tasks of triplet-based similarity measure in~\ref{subsec:locaTest} against state-of-the-art architectures of ResNet50 (R50), ResNext50 (RX50), VGG16 (V16) and GoogleNet (GN)~\cite{resnet, resnext, vgg16, googlenet}. The selected baselines are proved to be good image encoders cross tasks of high-resolution robot localization~\cite{vggUsage, googlenetUsage, kendall2017geometric, sattler2018benchmarking}, visual navigation~\cite{SPTM, maqueda2018event}, place recognition~\cite{NetVLAD,zhang2018collaborative}.
\subsection{Environment Setup}\label{subsec:Environment}
We discretize the environment to $d=0.25\rm m$ as mentioned in \ref{subsec:dataset} with maximum possible object instances $M=256$ which are further categorized to static, low-dynamic, and high dynamic objects by their object classes. We extract the scene graph from RGB images using a heuristic method. We build up our SeanNet using PyTorch~\cite{PyTorch}. We provide the code-base together with detailed instructions for collecting datasets and reproducing the results in \url{https://github.com/XiaoLiSean/Cognitive-Map/tree/codebase}. 
\subsubsection{Scene Graph Extraction}\label{subsubsec:sceneGraphGen}
\begin{figure}[H]
\centering
     \begin{subfigure}[b]{0.45\linewidth}
         \centering
        \includegraphics[width=0.8\linewidth]{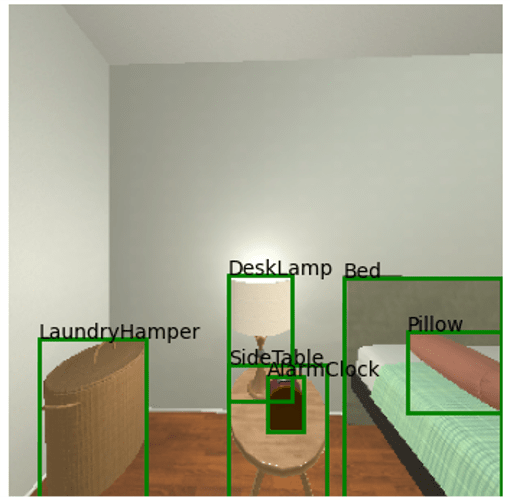}
         \label{fig:image_sgEx}
     \end{subfigure}
     \hspace{1em}
     \begin{subfigure}[b]{0.45\linewidth}
         \centering
        \includegraphics[width=0.9\linewidth]{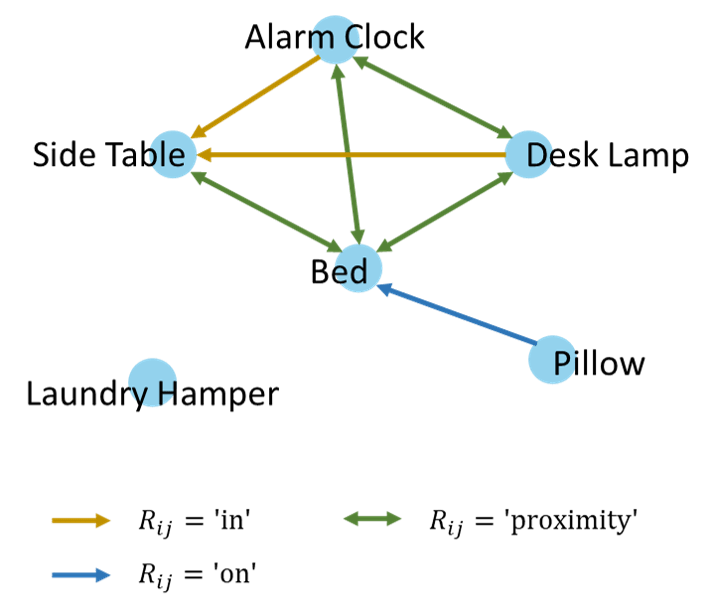}
         \label{fig:sg_sgEx}
     \end{subfigure}
\caption{Scene graph extraction.}  
\label{fig:sgEx}
\vspace{-1em}
\end{figure}

For each RGB image, we can abstract the semantic information using a scene graph as shown in Fig.~\ref{fig:sgEx}. The "proximity" edges are notated with double-direction arrows which imply the proximity relationship is mutual. As the four inter-object relationships in $\mathcal{R}$ are mutually exclusive, the "disjoint" edge connects the majorities of the objects and it's hidden in Fig.~\ref{fig:sgEx}.  

In real application scenario, we note that the scene graph generation from image can be implemented using existing methods~\cite{li2017scene,yang2018graph,xu2017scene}. And the topic of the scene graph generation is not in the scope of our problem. Therefore, in our experiment, we use a heuristic method for simplification. The AI2THOR environment provides information regarding if object $o_i$ is 'on' $o_j$. Meanwhile, for an object $o_i$, we can extract a 3D bounding box $B_i$ with eight corner points $b_{i,k}\in B_i,\; k=1,\dots,8$. Then, the other three inter-object relationships between $o_i$ and $o_j$ are defined as
\begin{equation*}
\left\{
    \begin{aligned}
       R_{ij} &= \text{'in' } &\text{if } B_i\subset B_j \\
       R_{ji} &= \text{'in' } &\text{if } B_j\subset B_i \\
       R_{ij} &= R_{ji} = \text{'proximity' } &\text{if } r(B_i,B_j) \leq r(B_i)+r(B_j) \\
       R_{ij} &= R_{ji} = \text{'disjoint' } &\text{otherwise }
    \end{aligned}
\right.
\end{equation*}
where $r(B_i)$ is the radius of the smallest sphere that bounds $B_i$ and $r(B_i,B_j)$ is the distance between the center of $B_i,B_j$. 
\subsubsection{Object Dynamics}\label{subsubsec:objDynamics}
We categorize the 125 types of objects into static, low dynamic, and high dynamic objects. The 16 structural object types such as sink and window are static. We assume 19 types of objects such as the coffee machine are low-dynamic and only change their pose locally. Specifically, the low dynamics object $o_i$ moves within a distance range of $r(B_i)$. The other object types such as chairs and desktops are highly dynamic and can move across the entire simulation environment. 

\begin{figure*}[t!]
\vspace{0.6em}
\begin{center}
\includegraphics[width=0.85\linewidth]{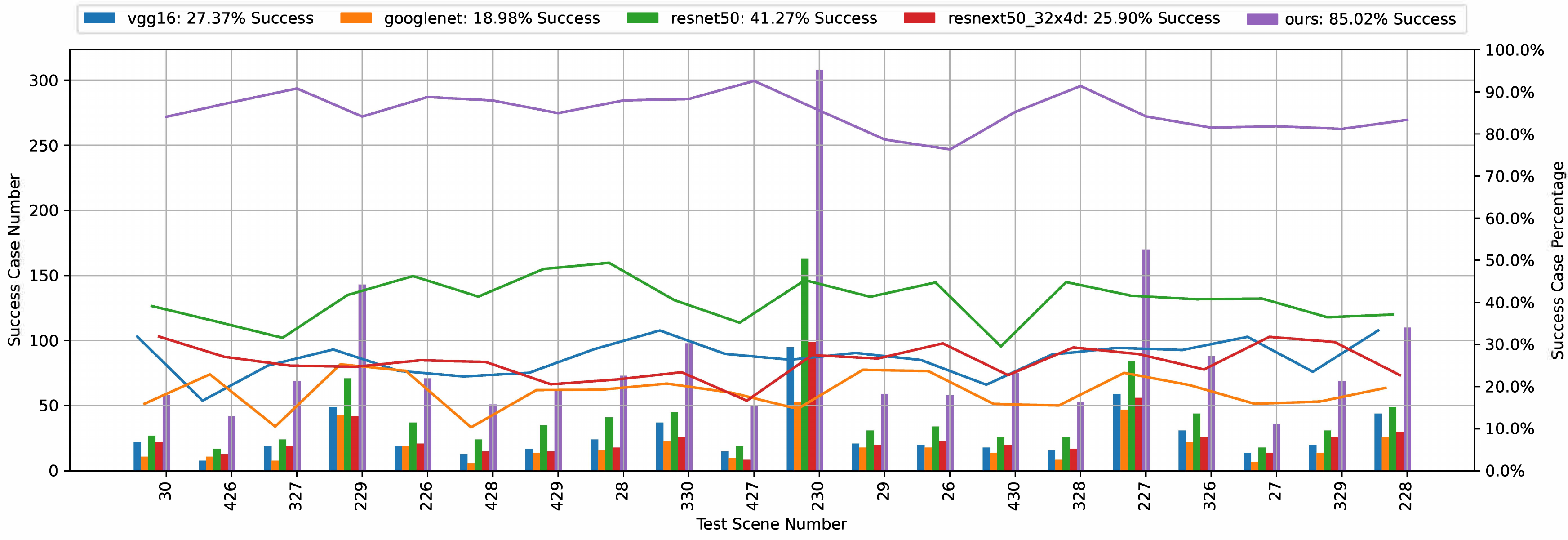}
\end{center}
\vspace{-1.0em}
\caption{Test statistics for triplet-based similarity measure.}
\label{fig:unitTest}
\vspace{-1em}
\end{figure*}

\subsection{Network Training Hyper-parameter}\label{subsec:hyperparam}
We train the networks in triplet dataset of 9.5k data points. The AI2Thor comprises four types of rooms (kitchen, living room, bedroom and bathroom) and each have 30 different instances. We select 21,4,5 room instances from each room types as training, validation and test rooms, respectively. To fully utilize the GPU capacity, the batch size of our baseline is 14 while that for VGG16, ResNet50, ResNext50, and GoogleNet are 19, 26, 20, 400 respectively. All networks are initialized with weights pre-trained on ImageNet~\cite{imagenet} and  trained using SGD optimizer of learning rate $lr=0.01$ with momentum of $\gamma = 0.9$. We deploy a learning rate scheduler for 60 epochs of training which decays $lr$ by a factor $\beta=0.7$ every 10 epochs. To mitigate over-fitting in training, a dropout rate of 0.2 is imposed on the GCNs in SeanNet. Meanwhile, we use a constant margin of $\alpha=0.1$ in triplet loss.   

\subsection{Similarity Measure Testing and Localization Threshold}\label{subsec:locaTest}

We evaluate the networks in the test dataset and consider $\mathcal{L}(f_A, f_P) > \mathcal{L}(f_A, f_N)$ as a success case. As shown in Fig.~\ref{fig:unitTest}, our architecture outperforms the benchmarks cross 20 test rooms. SeanNet can successfully transfer the learned similarity measure from the training rooms to unseen test rooms. The CNN benchmarks learned to match anchor and positive images under object dynamics, however, at the cost of sacrificing the ability to distinguish between $A$-$N$ images. Thus, the bottleneck of these CNN benchmarks is the $A$-$N$ ambiguity where the anchor and negative poses are close to each other but different in a fine-grained localization setting. In contrast, SeanNet can robustly match anchor-positive scenes under object dynamics, while still able to distinguish negative-anchor scenes given ambiguity.  

\vspace{-0.5em}
\begin{figure}[H]
\centering
     \begin{subfigure}[b]{0.44\linewidth}
         \centering
        \includegraphics[width=0.95\linewidth]{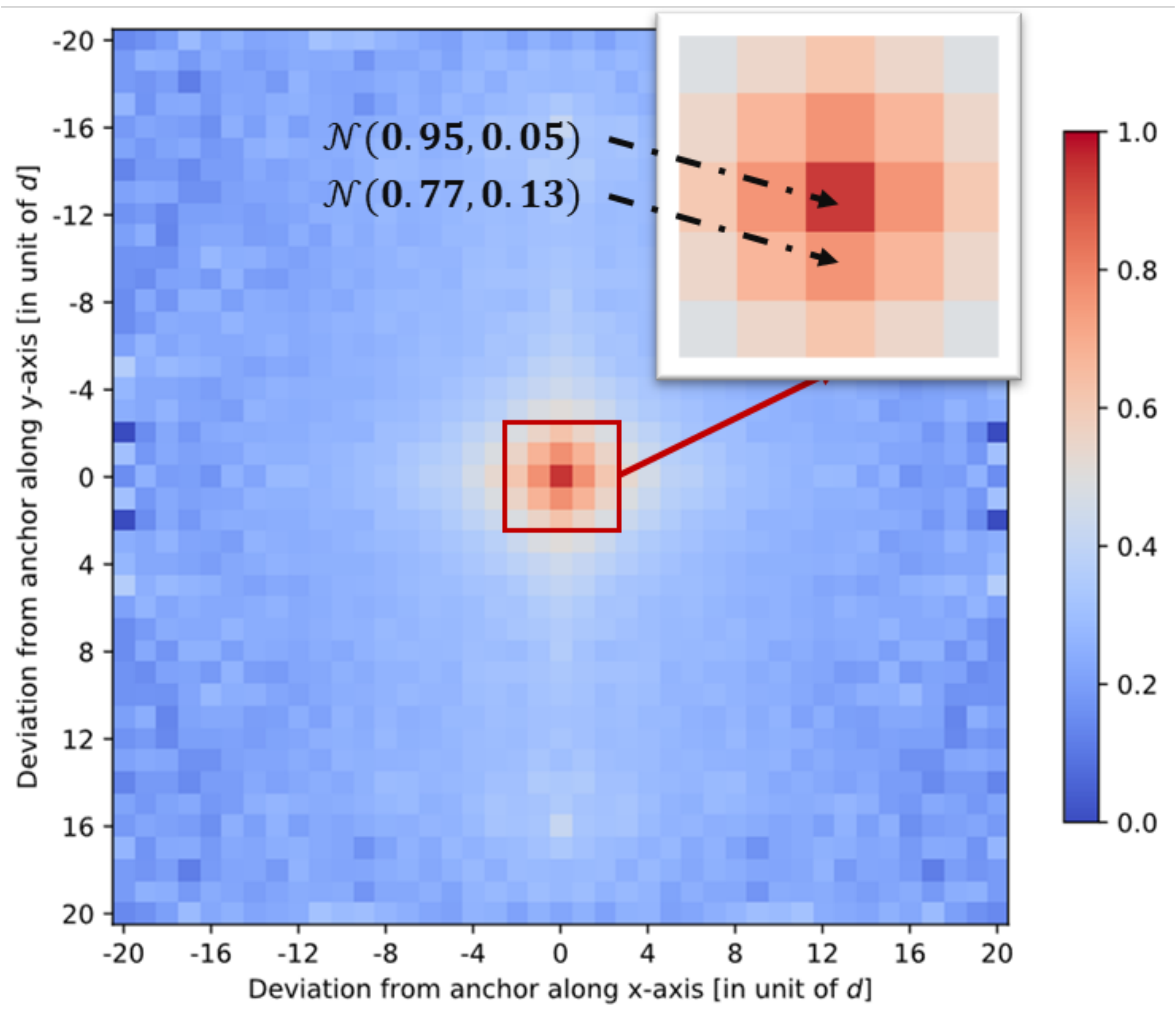}
     \end{subfigure}
     \hspace{2em}
     \begin{subfigure}[b]{0.44\linewidth}
         \centering
        \includegraphics[width=0.95\linewidth]{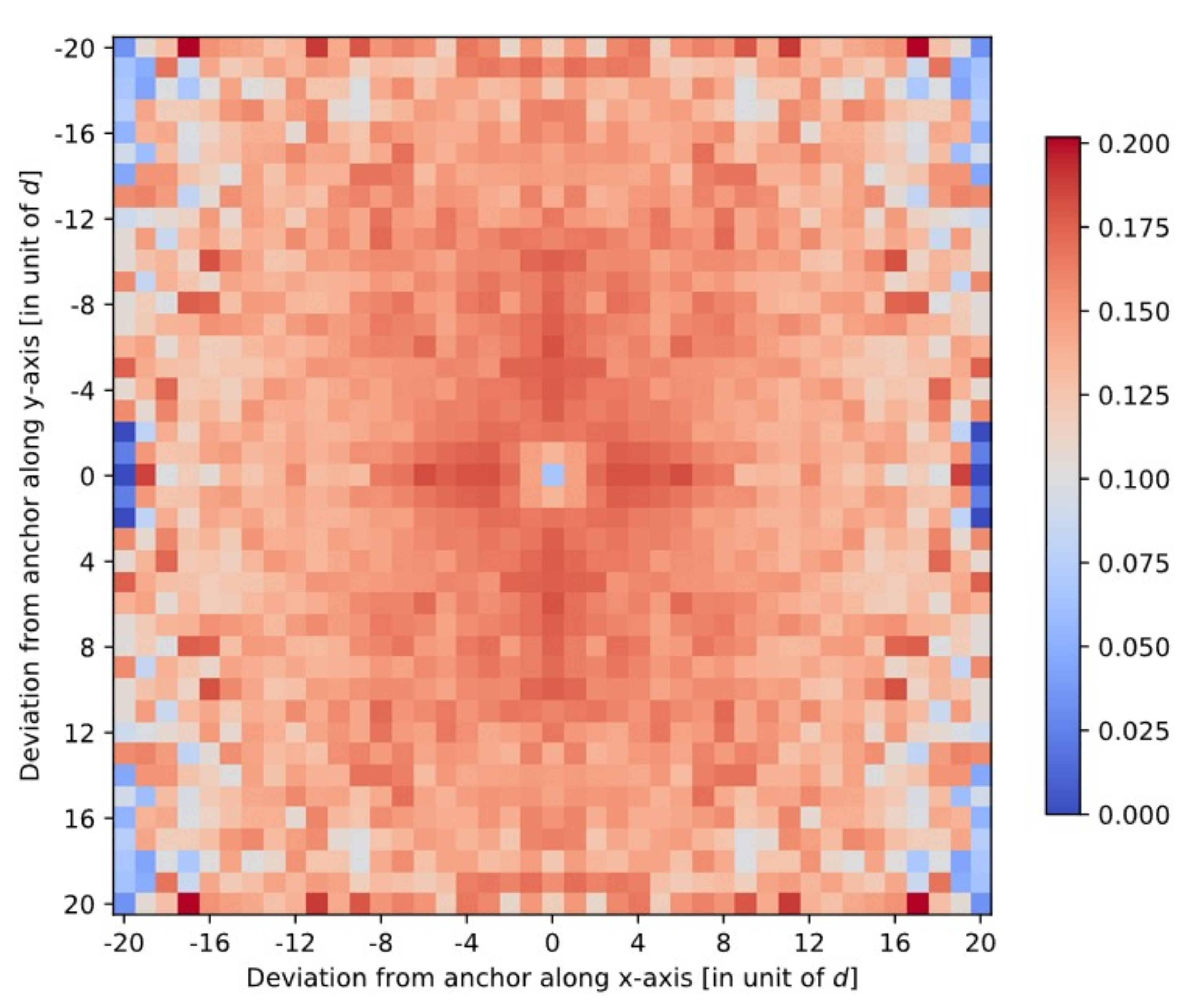}
     \end{subfigure}
\caption{Similarity distribution versus position deviation with \textbf{same robot heading angle}. (Left) Mean similarity value. (Right) Similarity standard deviation.}
\label{fig:threshold}
\vspace{-1em}
\end{figure}

As shown in Fig.~\ref{fig:threshold}, we collect 9.3k $(A,P)$ and $(A,N)$ pairs from validation dataset to visualize the distribution of their similarity values versus the position deviations along X-Y axis. We can see a clear peak mean similarity value in the center. For $(A,N)$ with different robot heading, the similarity is generally lower than 0.5, thus results are not presented. The pairs $(f_A,f_P)$ from the same pose in Fig.~\ref{fig:threshold} have similarity of $0.95\pm0.05$ while that for $(A,P)$ with one step deviation is $0.77\pm0.13$. Therefore, we can choose a threshold $\epsilon=0.90$ statistically such that one can conclude $(f_A, f_P)$ are feature observations from the same robot pose with $\mathcal{L}(f_A, f_P)>\epsilon$.

\begin{figure}[H]
\centering
     
         \centering
        \includegraphics[width=1\linewidth]{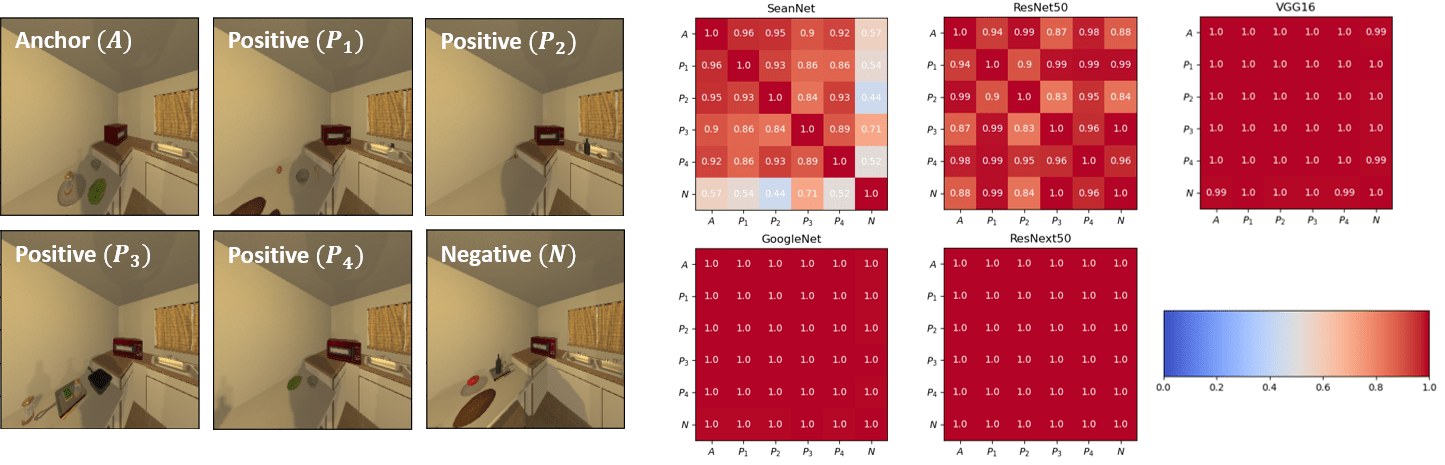}
     
     
     \vspace{-0.5em}
\caption{Similarity examples: $N$ is one step away from~$A$.}
\label{fig:tripletExamplesAppendix}
\end{figure}

\begin{figure}[H]
\centering
     \begin{subfigure}[b]{0.24\linewidth}
         \centering
        \includegraphics[width=1\linewidth]{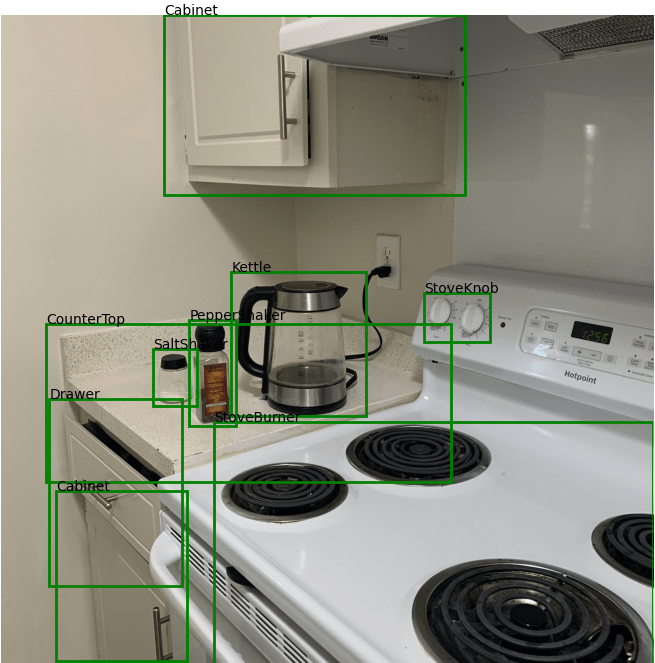}
        
        \vspace{-0.5em}
         \caption{Anchor $A$.}
     \end{subfigure}
     \begin{subfigure}[b]{0.24\linewidth}
         \centering
        \includegraphics[width=1\linewidth]{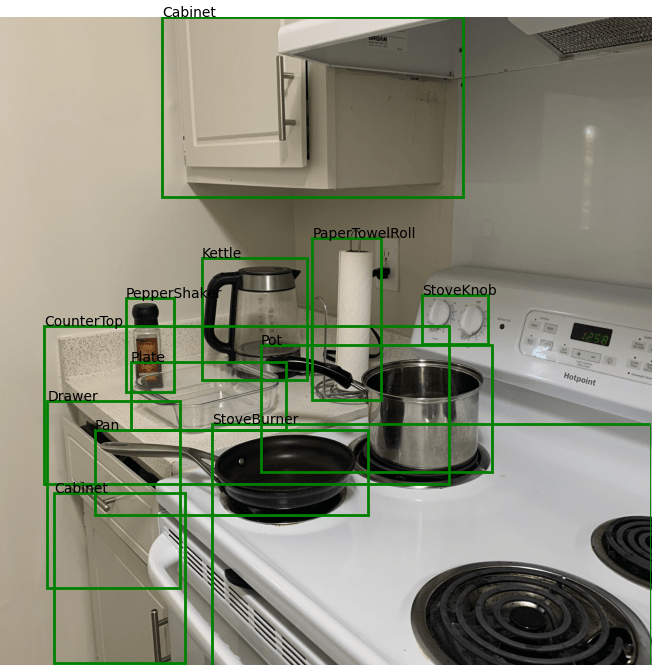}
        
        \vspace{-0.5em}
         \caption{Positive $P_1$.}
     \end{subfigure}
     \begin{subfigure}[b]{0.24\linewidth}
         \centering
        \includegraphics[width=1\linewidth]{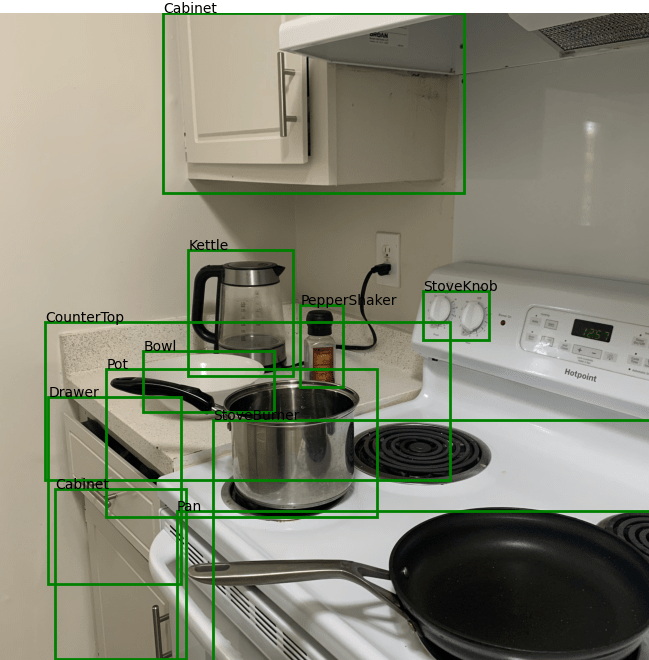}
        
        \vspace{-0.5em}
         \caption{Positive $P_2$.}
     \end{subfigure}
     \begin{subfigure}[b]{0.24\linewidth}
         \centering
        \includegraphics[width=1\linewidth]{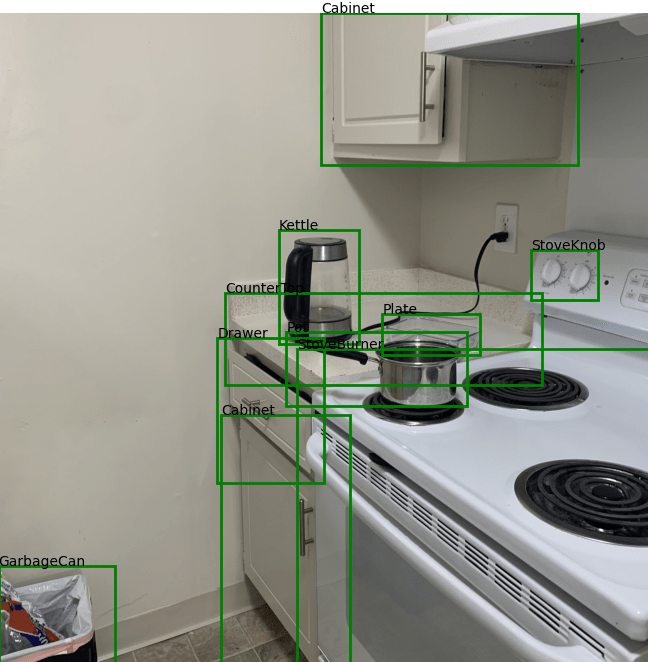}
        
        \vspace{-0.5em}
         \caption{Negative $N$.}
     \end{subfigure}  
     
     \begin{subfigure}[b]{1\linewidth}
         \centering
        \includegraphics[width=1\linewidth]{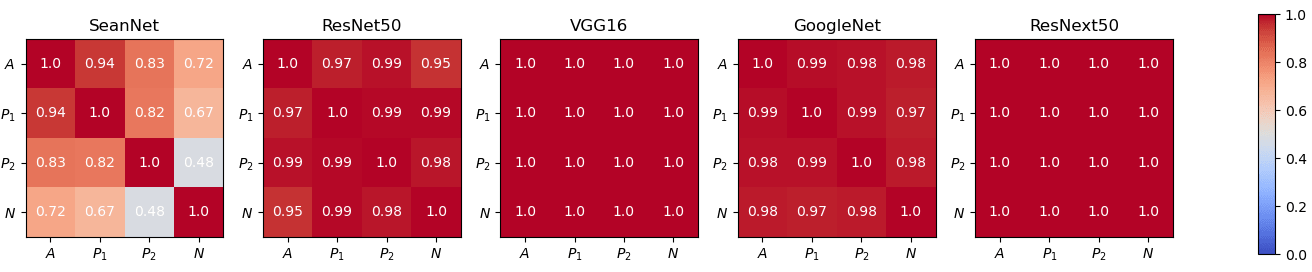}
     \end{subfigure}
     
     \vspace{-0.5em}
\caption{Real-world similarity examples}
\label{fig:tripletExamplesReal2}
\end{figure}

We provide examples with $A$-$N$ ambiguity akin to Fig.~\ref{fig:triplet_example} from both test rooms and real world. As shown in Fig.~\ref{fig:tripletExamplesAppendix}, each entry in the heat-map shows the pairwise similarity score between the features (e.g. $\mathcal{S}(f_A,f_{P_1})$, $\mathcal{S}(f_{P_2},f_{N})$) using the proposed SeanNet and benchmarks. The SeanNet outperforms the benchmarks by successfully generating results of $\mathcal{S}(f_A,f_{P_i})>\mathcal{S}(f_A,f_{N}),\;i=1,2,3,4$ while he second best ResNet50 fails in majorities of the data points. Meanwhile, for majorities of the triplets, the SeanNet produces similarity gaps $\mathcal{S}(f_A,f_{P_i})-\mathcal{S}(f_A,f_{N})>\alpha=0.1$ which originates from the proposed triplet loss and guarantees a good localization performance. As shown in Fig.~\ref{fig:tripletExamplesReal2}, we manually label the object bounding box and scene graphs. The aforementioned phenomenon hold for these real world examples. 

\vspace{-0.5em}

        
        
        
        
     
     

\vspace{-1.5em}

%% file: tex/application.tex
\begin{figure*}
\vspace{0.6em}
\begin{center}
\includegraphics[width=0.95\textwidth]{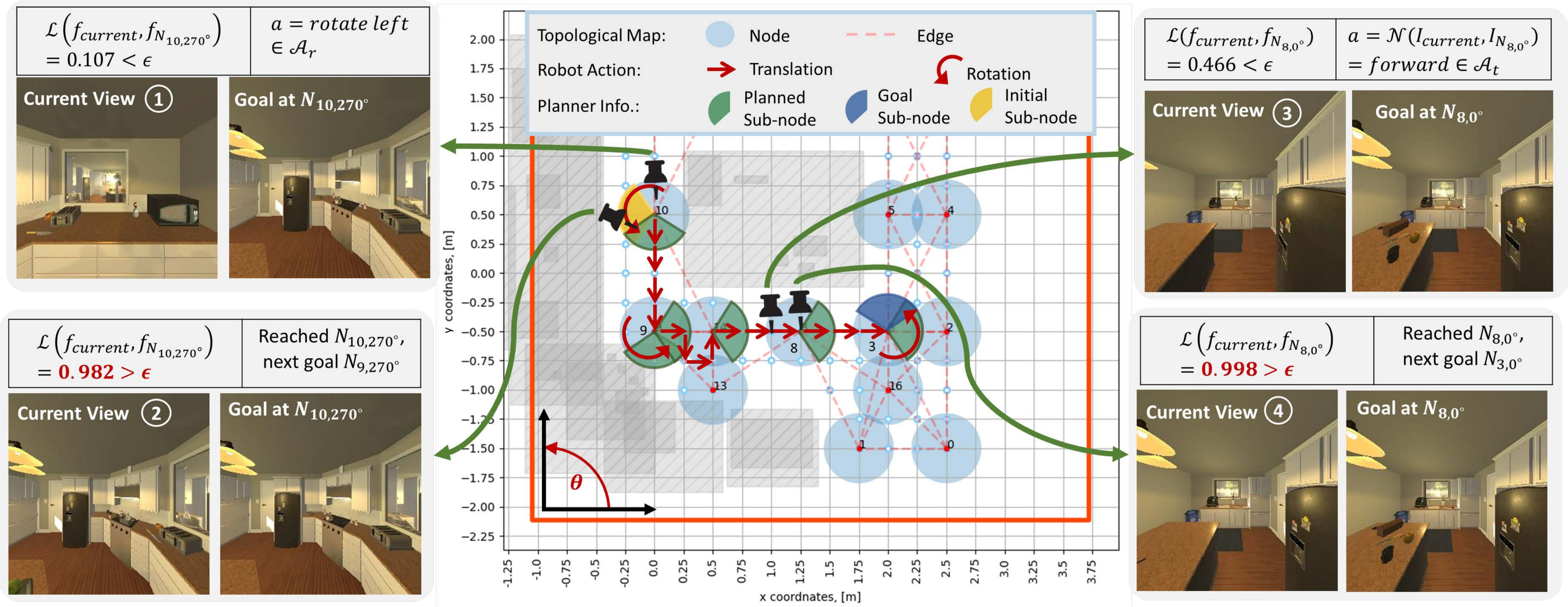}
\end{center}
\vspace{-0.5em}
\caption{\textbf{Sequential navigation and localization.} From view 1-2 where the view 1 is previously localized to a sub-node,  the rotation edge connects the current sub-node with the goal sub-node and generates the counter-clockwise rotation action. From view 3-4, the visual navigator generates a rightward action command. A demo video is avaible in \url{https://github.com/XiaoLiSean/Cognitive-Map/tree/codebase}.}
\label{fig:SequentialExample}
\vspace{-1.5em}
\end{figure*}

To demonstrate how the SeanNet can be used as localizer in practice, we define a visual navigation task under object dynamics with topological mapping \cite{kuipers2000spatial} which has been long studied and can enlarge the visual navigation range by a large margin \cite{SPTM, fraundorfer2007topological}. We use a visual navigator to navigate the robot within the space $\mathcal{P}$ by taking action $a_k \in \mathcal{A}_r\cup\mathcal{A}_t$ where $\mathcal{A}_r=\{\text{rotate left }90^{\circ},\;\text{rotate right }90^{\circ}\}$ and $\mathcal{A}_t=\{\text{forward } d,\;\text{leftward } d,\;\text{rightward } d\}$. Given arbitrary initial pose $A\in\mathcal{P}$ and target poses $B\in\mathcal{P}$, we consider a visual navigation process $\{a_k\}_{k=1}^{n}$ which navigates the robot from $A$ to $B$. We aim for monitoring the navigation and terminating the navigation sequence at the target $B$ with the SeanNet-based localizer.

\subsection{Topological Map}\label{subsec:topo_map}
Inspired by~\cite{fraundorfer2007topological, kuipers2000spatial}, we represent the discrete space $\mathcal{P}$ with a topological graph $\mathcal{TG}(N,E)$ composed by nodes $N$ and edges $E$ as in Fig.~\ref{fig:SequentialExample}. Each node $N_i$ encodes a topological place in the environment and contains four sub-nodes $N_i=\{N_{i,k}\}, k=0,90,180,270^{\circ}$ that represent four robot headings. The \textbf{translation edge} $E_{ij}$ connecting adjacent nodes $N_i$, $N_j$ encodes trasversability and stores Manhattan distance as cost between them. The sub-nodes $\{N_{i,k}\}$ in the same node $N_i$ are interconnected with \textbf{rotation edges} and stores features $f_{N_{i,k}}=\{I, \mathcal{SG}\}$ from robot observation during mapping phase.

\subsection{Visual Navigation Pipeline}\label{subsec:visual_navi}
Given $\mathcal{TG}(N,E)$, we define the visual navigation task as navigation from one sub-node $N_{i,k_1}$ to arbitrary goal sub-node $N_{j,k_2}$. Given the graph nature of $\mathcal{TG}$ and the cost metrics stored in the edges, we use Dijkstra algorithm for planning a shortest path from $N_{i,k_1}$ to $N_{j,k_2}$ of which each individual segment contains two neighboring sub-nodes in the map. Using similar method proposed in~\cite{SPTM}, we train a visual navigator $\mathcal{N}(I_A,I_B)$ with VGG16 that predicts the desired translation action $a\in\mathcal{A}_t$ given current and target image (stored in the sub-node) $(I_A,I_B)$. Eventually, the robot can navigate cross each segment in the path by either taking structurally-informed rotation action or action from the visual navigator. The similarity-based localization in \ref{subsec:locaTest} determines if the current navigation segment is completed, thus update the navigation task with next segment. A demonstration of this navigation process is shown in Fig.\ref{fig:SequentialExample}.

\begin{table}[htbp]
  \centering
  \caption{Navigation test: within neighbour sub-nodes (N) and cross arbitrary (A) sub-nodes.}
    \begin{tabular}{c|c|c|c|c|c}
    \hline\hline
    \multirow{3}[0]{*}{} & Total & Success  & \multicolumn{3}{c}{Failure rate (\%)} \\\cline{4-6}
          & case & rate (\%) & \scriptsize{Navigation} & \scriptsize{Localization} & \scriptsize{Collision} \\\cline{2-6}
          & \scriptsize{N / A}     & \scriptsize{N / A}     & \scriptsize{N / A}     & \scriptsize{N / A}     & \scriptsize{N / A} \\\hline\hline
    Ours  & & \textbf{80.2} /	\textbf{65.2} &	2.4 /	6.2 &	\textbf{13.2} /	\textbf{21.7} &	4.1 /	6.8   \\
    R50 & \scriptsize{5476} & 29.0 /	3.4 &	\textbf{0.3} /	0.5 &	70.2 /	95.2 &	\textbf{0.4} /	0.9   \\
    RX50 &  /  & 26.0 /	14.5 &	0.6 /	\textbf{0.48} &	72.8 /	84.7 &	0.7 /	\textbf{0.3}   \\
    V16 & \scriptsize{204136} & 62.3 /	20.1 &	0.7 /	1.8 &	35.5 /	74.2 &	1.5 /	3.9   \\
    GN & & 69.4 /	30.6 &	1.8 /	3.3 &	25.5 /	58.8 &	3.3 /	7.3   \\\hline\hline
    \end{tabular}%
  \label{tab:visualNaviData}
  \vspace{-1.5em}
\end{table}

We conduct a navigation test as shown in Table~\ref{tab:visualNaviData} using the same visual navigator and different localizers, i.e., with SeanNet and other baselines. We consider a test trial \textit{failure due to localization} if the localizer fails to terminating navigation while the current goal sub-node is actually reached or if there is a false-positive localization signal. If the robot fails to reach the target within $k_{max}=12$ steps without a failure in localization, we consider this test case \textit{navigation failure}. Meanwhile, the robot can fail in a navigation test due to \textit{collision} with the environment. Our network has the best averaged success rate across 20 test rooms which is due to the localizer based on SeanNet has a lower localization failure rate.

%% file: tex/conclusion.tex
We proposed a novel network architecture that can incorporate visual and semantic information to learn a vector embedding for scene observation with object dynamics, whereby a network-based similarity measure has been developed. Based on the SeanNet trained with the proposed cascaded constrastive learning scheme, we have shown that the similarity-based localization function is robust to object dynamics and can successfully monitor and terminate the visual navigation process. For the future work, we are interested in adapting our similarity measure to outdoor localization facing changing perceptual conditions.